\documentclass[12pt]{article} 
\usepackage{xcolor}
\usepackage{arabtex}
\usepackage{amsmath}
\usepackage{coling2018}
\usepackage{ragged2e}
\usepackage{url}
\usepackage{hhline}
\usepackage{booktabs}
\usepackage{float}
\usepackage{graphicx,subcaption}
\usepackage{smartdiagram}
\usepackage{pgf-pie}
\usepackage{enumitem}
\usepackage[para,online,flushleft]{threeparttable}

\usepackage[T1]{fontenc}
\usepackage{textcomp}
\usepackage{lmodern}
\usepackage{silence}
\usepackage{caption} 

\usepackage{colortbl}
\usepackage[stable]{footmisc}
\usepackage{cp1256,utf8}
\setcode{utf8}


\definecolor{Kalkola}{HTML}{EE171F} %

\date{}

\definecolor{lightgray}{rgb}{0.95,0.95,0.95}

\begin{document}

\title{Which one Performs Better? Wav2Vec or Whisper? \\ Applying both in Badini Kurdish Speech to Text (BKSTT)}

\author{
	\begin{tabular}[t]{c}
		Renas Adnan and Hossein Hassani\\
		\textnormal{University of Kurdistan Hewl\^er}\\
		\textnormal{Kurdistan Region - Iraq}\\
		{\tt {\{renas.adnan, hosseinh}\}@ukh.edu.krd}
	\end{tabular}
}

\maketitle
\begin{abstract}
Speech-to-text (STT) systems have a wide range of applications. They are available in many languages, albeit at different quality levels. Although Kurdish is considered a less-resourced language from a processing perspective, SST is available for some of the Kurdish dialects, for instance, Sorani (Central Kurdish). However, that is not applied to other Kurdish dialects, Badini and Hawrami, for example. This research is an attempt to address this gap. Bandin, approximately, has two million speakers, and STT systems can help their community use mobile and computer-based technologies while giving their dialect more global visibility. We aim to create a language model based on Badini's speech and evaluate its performance. To cover a conversational aspect, have a proper confidence level of grammatical accuracy, and ready transcriptions, we chose Badini kids' stories, eight books including 78 stories, as the textual input. Six narrators narrated the books, which resulted in approximately 17 hours of recording. We cleaned, segmented, and tokenized the input. The preprocessing produced nearly 15 hours of speech, including 19193 segments and 25221 words. We used Wav2Vec2-Large-XLSR-53 and Whisper-small to develop the language models. The experiments indicate that the transcriptions process based on the Wav2Vec2-Large-XLSR-53 model provides a significantly more accurate and readable output than the Whisper-small model, with 90.38\% and 65.45\% readability, and 82.67\% and 53.17\% accuracy, respectively.
\end{abstract}

\section{Introduction}

Human-machine verbal communication becomes inseparable from daily activities \cite{alharbi2021asr}. In response to this growth, Speech-to-Text (STT) receives increasing attention in Natural Language Processing (NLP), both in research and application development. The terms Speech-to-Text (STT) and Automatic-Speech-Recognition (ASR) are often used interchangeably in academic literature and industrial terminology \cite{singh2023stt,gyulyustan2024stt}. While ASR is more popular in technical contexts, STT is commonly used in consumer-focused applications. In this paper, we use STT for its popularity.

While STT has made significant advancements in well-studied languages, low-resource languages are still far behind. Spoken by approximately 1.5 to 1.7 million people in the Kurdistan Region of Iraq (KRI), particularly in Duhok governorate \cite{othman2022population}, Badini is a Kurdish dialect that currently lacks the necessary resources required for many NLP tasks, particularly STT. This research aims to reduce this gap by creating a speech corpus and developing a language model for this dialect.

We develop the language models based on two frameworks, Wav2Vec2 and Whisper, to assess their suitability and performance for use in various applications. This approach is in line with the latest findings regarding STT. For example, \cite{bhogale2024pseudo} proposes `pseudo-labeling' to expand datasets by generating machine-assisted transcriptions, thereby supplementing limited data without exhaustive human labor. Similarly, \cite{getman2024fine} stated the effectiveness of fine-tuning multilingual models, such as Wav2Vec 2.0, when minimal data for the target language is available. Cross-lingual transfer learning has also been used to improve STT for low-resource languages. For example,  \cite{li2024cross} showed that self-supervised learning (SSL) techniques, such as those in HUBERT and Wav2Vec, will provide better ways to train SST models using unlabeled data.

The rest of this paper is organized as follows. Section 2 provides the related work, Section 3 describes the research method, Section 4 presents the results, and Section 5 concludes the research and provides future expansion areas.

\section{Related Work}

Transformer models have revolutionized natural language processing. They have the potential to capture the long-range dependencies present in speech, where traditional models, such as Recurrent Neural Network (RNN), fail to perform non-casual forward computations \cite{lu2020exploring}. According to \newcite{lu2020exploring}, the self-attention mechanism of transformers allows wide and efficient contextual understanding, and they can easily capture the temporal dependency at each position of the input sequence. Transformer models have become part of real-world STT systems, showing significant improvements in their efficiency \cite{orken2022study}. 

In recent years, large-scale STT foundation models such as Whisper\cite{radford2022Whisper} have become increasingly popular and available, further lowering the entry barrier for building or extending STT systems for currently unsupported or under-supported languages. The Whisper models utilize Transformer technology and can be used in most languages. It has been trained over 680,000 hours of labeled audio and can work effectively even without significant extra training \cite{radford2022Whisper}. Unlike other models, Whisper works reasonably well in transcribing under-resourced languages \cite{do2023fine}.

To improve performance on low-resource (STT) tasks, several works have fine-tuned Whisper on specific datasets. As an example, Whisper-Tiny fine-tuned on Labeled Oral Reading Assessment (LoRA), achieved 38.49 percent Word Error Rate (WER) improvement over zero-shot settings \cite{do2023fine}. Furthermore, beam search algorithm modification along with adding Filter-Ends decoding techniques reduced the average WER by another 2.26 percent, which demonstrates that end-to-end optimized decoding techniques perform better in low-resource STT workloads \cite{jain2023child}.

However, rapidly fine-tuning Whisper to a different language or domain from its inceptive one is not always successful, neither in terms of data preparation nor the learning process. Training STT models typically requires large-scale transcribed corpora, which are not available for many low-resource languages. When using Whisper, which is fine-tuned from pre-trained checkpoints of another language, it still underperforms a deep monophony conventional triphone system \cite{ma2023adapting}. According to \newcite{ma2023adapting}, Whisper tends to omit disfluencies and hesitations in its output, which may be harmful to accurate candidate assessment in a spoken language assessment application. Applying either soft-prompting or fine-tuning enables Whisper to generate decoding output that could be better tuned for use in spoken language assessment. 

\newcite{Das2021MultiDialect} addressed the challenge of the variation of English dialect by proposing an ensemble of transformer-based models that were trained on Australian,  American, Canadian, and British English. They achieved the best value with 4.74\% of WER reduction over the baseline. This strategy showed the potential of attention-based models in generalizing effectively to various dialects while maintaining performance levels. For developing and enhancing accuracy, some techniques for extracting features, such as the Mel Frequency Cepstral Coefficients (MFCCs) were applied. The MFCC provided an efficient and compressed method of representing speech signals and hence enabled the process of differences in speech among speakers and noise \cite{abdalla2010wavelet}. The techniques that this research applied were important in cleaning and normalizing sound signals prior to their presentation in speech-to-text systems.

To summarize, earlier research used a mixture of machine learning approaches that have led to today's stage, with initial STT systems being based primarily on statistical models like HMM\textbackslash GMM and N-gram language models, which were assisted by feature extraction methods like MFCC.  With the advent of deep learning, modern methods like DNNs, RNNs, and, more recently, transformer-based models have considerably enhanced transcription quality, especially in noisy environments and low-resource languages like Kurdish. Further, end-to-end models and transfer learning approaches have simplified pipeline complexity and increased performance. For a high-quality STT system, high-quality recordings are required. However, this approach is expensive and requires continuous effort to collect and annotate clean speech data. Recently, the best-performing models have shifted to neural network-based architectures. Whisper and Wav2Vec2 are examples of this, following an end-to-end transformer-based design.

\section{Method}
The following sections describe the approach of the project. They explain data collection and preparation methods, setting up the environment to train the model, and enhancing pre-trained models like Wav2Vec2, WavLSTM, and XLS-R. We verify how well the model performs using quality metrics. Finally, we test the model in a real-life environment to check its clarity across various sound backgrounds.

\subsection{Data Collection}
The goal is to collect approximately 10 to 20 hours of Badini speech data. We select a text collection for which we can obtain permission to publicize its result models because our objective is to contribute to data augmentation for Kurdish NLP. We recruit narrators to narrate the selected texts, and we consider the following in this process:

\begin{itemize}[itemsep=0pt, topsep=0pt, partopsep=0pt, parsep=0pt]
    \item Finding fluent native and non-native Badini speakers. 
    \item Evaluating the quality of narrations, particularly pronunciations. The narrators are given the same sample stories to read. The recordings are checked for clarity, pronunciation correctness, and to double-check the environment's noise level. The selected narrators are asked to resolve the issues raised during the testing.
    \item Preparing professional devices for the recording process. The narrators post their recordings on a shared Mega Drive folder daily. The records are checked for background noises, pronunciation accuracy, balanced speech rate, and correct segmentation of utterances. If a record was flawed (e.g., mispronunciations, background noise, or unnatural pace), it is sent back for re-recording. The approved items are kept in a properly structured dataset.
\end{itemize}

\subsection{Preporcessing}
We use noise reduction methods in Adobe Audition (2022), which include the following steps:

\begin{center}
   \begin{itemize}[itemsep=0pt, topsep=0pt, partopsep=0pt, parsep=0pt]
        \item Import the recorded audio files into Adobe Audition.
        \item Select the audio to be activated in the timeline area, then find a noisy area to capture as a sample for processing and removing, as presented in Figure~\ref{capturenoise}.
        \item We reduce the noise by applying the captured noise profile to separate and remove extraneous sounds without disturbing the speaker's natural voice (see Figure~\ref{NoiseReduction}). The rendering produces a zero-line wave indicating the success of the noise removal function (see Figure~\ref{zeronoise}).
        \item We enhance the voice by eliminating the remaining noise. A script helps in applying a filter manually to completely separate the noise from the actual voice. Figure~\ref{before-noise} demonstrates the wave before applying the noise reduction, and Figure~\ref{after-noise} shows it after applying the filter. As the figures illustrate, the blue wave that represents the actual voice is effectively separated from the noise in the highlighted area.
        \item  The Diagnostics panel in Adobe Audition can be used for repairing audio issues. It allows us as editors to detect and fix problems such as silence, clipping, and noise after applying effects automatically (see Figure~\ref{diagnostics}).
        \item We batch process and export the resulting data using the Batch Processing Panel (see Figure~\ref{batch_processing}) and store the processed files (see Figure~\ref{Export_option}).  
    \end{itemize}      
\end{center}


\subsection{Segmentation and Alignment}
We either manually segment the data by demarcating the records at the sentence level according to the text, or automatically by scanning the audio to let the function select the proper sentence (see Figure~\ref{labling_names}). Each produced item is labeled by combining three indicators: narrator name, story title, and marker label, for example: \textit{San+NewDay+Marker01}.  
\subsection{Model Development}
To develop the models, we follow standard guidelines, utilize pre-trained models accessible in the Hugging Face Transformers library, and use two state-of-the-art multilingual speech recognition architectures: Wav2Vec2-XLSR-53 and Whisper-small. For the infrastructure, we use Google Colab Pro. Since the expectation is to collect between 10 and 20 hours of speech data, a large amount of RAM and extended storage capacity are necessary. To accelerate training speed and avoid session timeouts, we rely on the Tesla (T4) GPU, which provides balanced performance and cost.

\subsection{Evaluation}
The evaluation process is based on standard STT evaluation metrics, Word Error Rate (WER) and Character Error Rate (CER). A custom function calculates WER and CER using the model predictions and reference transcriptions. To maintain alignment between the model's output (model transcription) and the reference transcription, the function inserts a padding token ID for the not-found elements (ignored labels or unknown terms) in the output. Formulae~\ref{eq:accuracy_wer} and ~\ref{eq:accuracy_cer} show model accuracy for WER and CSE, respectively.

\begin{equation}
    {WER\:\:Accuracy} = (1 - {WER}) \times 100
    \label{eq:accuracy_wer}
\end{equation}
where \textit{WER} is the Word Error Rate

\begin{equation}
    {CER\:\:Accuracy} = (1 - {CER}) \times 100
    \label{eq:accuracy_cer}
\end{equation}
where \textit{CER} is the Character Error Rate

\section{Results and Discussion}

This section presents the research results and discusses the findings.

\subsection{Collected Data}
\begin{figure}[!ht]
        \centerline{\includegraphics[scale=0.08]{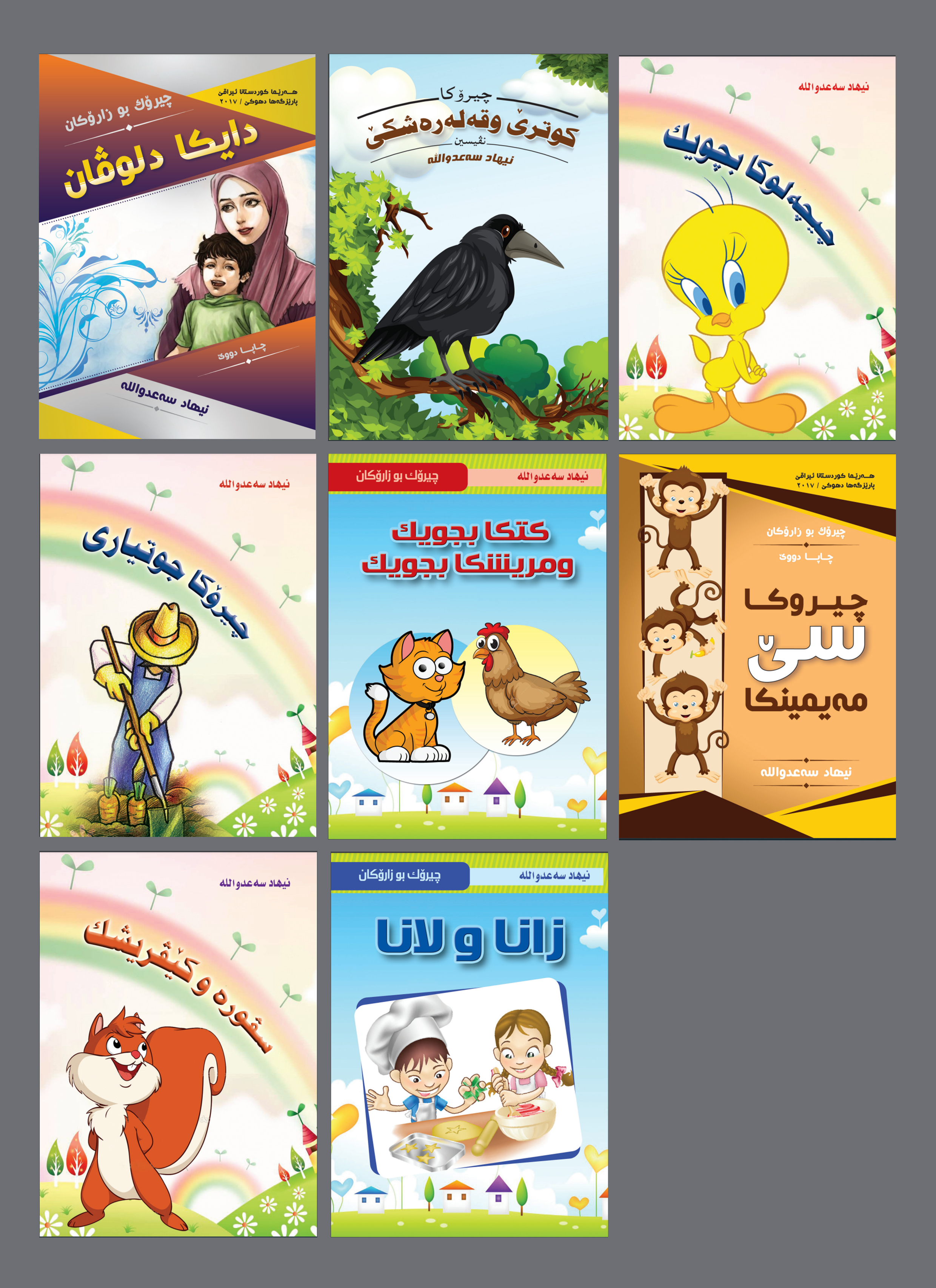}}
        \caption{Books selected for narrations}
        \label{Books}
\end{figure}

We shortlisted 78 stories extracted from eight books (see Figure~\ref{Books}) and recruited six Badini narrators, five female and one male. The selection was based on 111 Badini fiction stories from 12 books written for children between 6 and 12 years old.

\begin{figure}[!ht]
\centering
    \begin{subfigure}{.49\textwidth}
    \centering
        \includegraphics[width=0.6\textwidth]{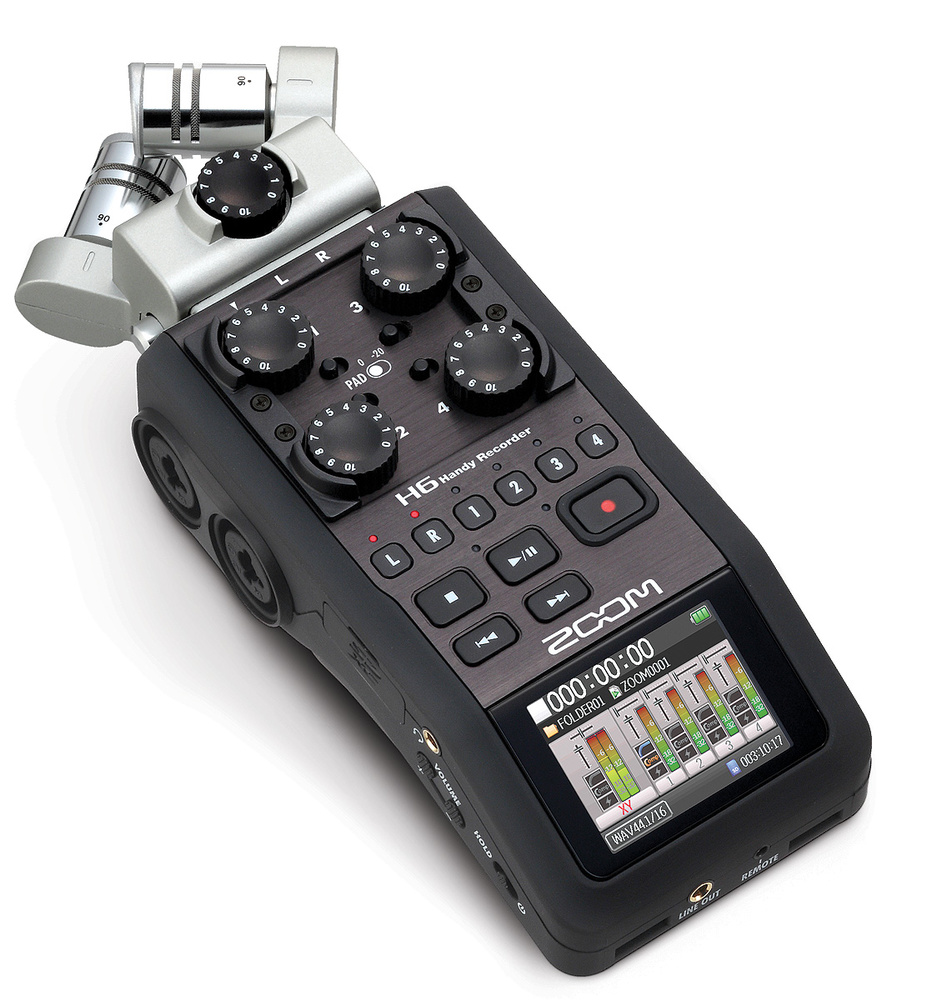}
        \caption{Recorder:Zoom H6Track portable}
        \label{zoomh}
    \end{subfigure}
    \begin{subfigure}{.49\textwidth}
    \centering
        \includegraphics[width=0.83\textwidth]{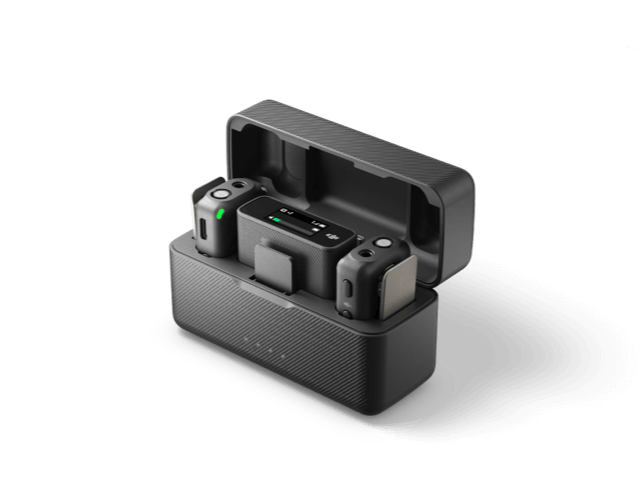}
        \caption{Microphone: DJI Mic 2 }
        \label{mic2}
    \end{subfigure}
    \caption{Recording devices}
    \label{fig:dev}
\end{figure}

Narrators received professional devices (see Figure~\ref{fig:dev}) to record their narrations and used a pop filter to reduce breathing sounds and loud pops. Inexperienced narrators also received brief training to ensure optimal use of the devices. The narrators recorded their narration in an environment of low background noise and echo. The audio recording used a 44.1 kHz sampling frequency, a monophonic channel, 32-bit depth, and a WAV file format.

The recording process produced approximately 30,000 audio tracks. We checked the quality of the records and found quality issues in about 0.14\% of them. This resulted in asking the two narrators who faced the problem to re-record those parts. 

\subsection{Segmentation and speech-text alignment}
The PDF formats of the books were converted to text. A Kurdish linguist assessed the text, ensuring the correct transformation and orthographic accuracy afterward. Below, we briefly describe the segmentation and alignment processes for the target models.

Wav2Vec2 models use the Connectionist Temporal Classification (CTC) algorithm, which considers timing in sequence-to-sequence problems. Since a pretrained tokenizer was not available for Badini, we created a custom character-level vocabulary for each model, based on the related requirements. That reduces vocabulary size and improves model generalizability for low-resource languages. We also used special tokens such as <pad>, <unk>, and a word delimiter symbol (|) (see Table~\ref{tab:special_tokens_vocab} to ensure compatibility with the CTC decoding process. Each character is mapped to an index after being saved in JSON format, which helps the tokenizer to `understand' the raw audio outputs. This approach allowed us to achieve a precise alignment between the speech signals and their corresponding written transcriptions.

We created a custom vocabulary to enhance the training process in both models. Since no pretrained Badini tokenizer exists currently, creating a character-level vocabulary from the dataset, which contains all Kurdish character IDs, enables proper alignment between audio and text under the CTC framework. It allows the model to accurately tokenize and decode Kurdish-specific phonemes, improving transcription accuracy, especially WER.

Whisper works based on a Byte Pair Encoding (BPE) tokenizer, which functions at the sub-word level. Because it is pre-trained on a variety of languages, we did not create a custom vocabulary manually. Instead, to manage different elements within the text, we depended on special tokens listed in Table~\ref{special_tokens-WS}.

\begin{table}[!ht]
\centering
\caption{Special Tokens Used in the Kurdish Vocabulary for \texttt{Wav2Vec2-Large-XLSR-53}}
\label{tab:special_tokens_vocab}
\begin{tabular}{|l|l|l|} 
    \hline
    \cellcolor{lightgray}\textbf{Token} & 
    \cellcolor{lightgray}\textbf{Index} & 
    \cellcolor{lightgray}\textbf{Purpose} \\
    \hline
    \texttt{<pad>} &  0 &  Pads sequences to the same length during batching. \\ 
    \hline
    \texttt{<s>} &  1 &  Marks the start of a sequence. \\ 
    \hline
    \texttt{</s>} &  2 &  Marks the end of a sequence. \\ 
    \hline
    \texttt{<unk>} &  3 &  Handles unknown characters not found in the vocabulary. \\ 
    \hline
    \texttt{|} &  4 &  Used as a word separator replacing regular spaces. \\ 
    \hline
    \texttt{" "} (space) &  5 &  Used to preserve spacing between words during decoding. \\ 
    \hline
\end{tabular}
\end{table}

\begin{table}[!ht]
\centering
\caption{Special Tokens Used in the Kurdish Vocabulary for \texttt{Whisper-small}}
\label{special_tokens-WS}
\begin{tabular}{|l|l|l|} 
    \hline
    \cellcolor{lightgray}\textbf{Token} & 
    \cellcolor{lightgray}\textbf{ID} & 
    \cellcolor{lightgray}\textbf{Description} \\
    \hline
    \texttt{<|startoftranscript|>} &  50258 &  Indicates the start of a transcription. \\ 
    \hline
    \texttt{Language Tokens} &  50259--50357 &  Specify the language of the input. \\ 
    \hline
    \texttt{<|translate|>} &  50358 &  Signals a translation task. \\ 
    \hline
    \texttt{<|transcribe|>} &  50359 &  Signals a transcription task. \\ 
    \hline
    \texttt{<|nospeech|>} &  50362 &  Indicates that the segment contains no speech. \\ 
    \hline
    \texttt{<|notimestamps|>} &  50363 &  Used to suppress timestamp generation. \\ 
    \hline
    \texttt{<|timestamp-begin|>} &  50364 &  Marks the beginning of timestamp tokens. \\ 
    \hline
\end{tabular}
\end{table}

\begin{figure}[!ht]
    {\centerline{\includegraphics[scale=0.5]{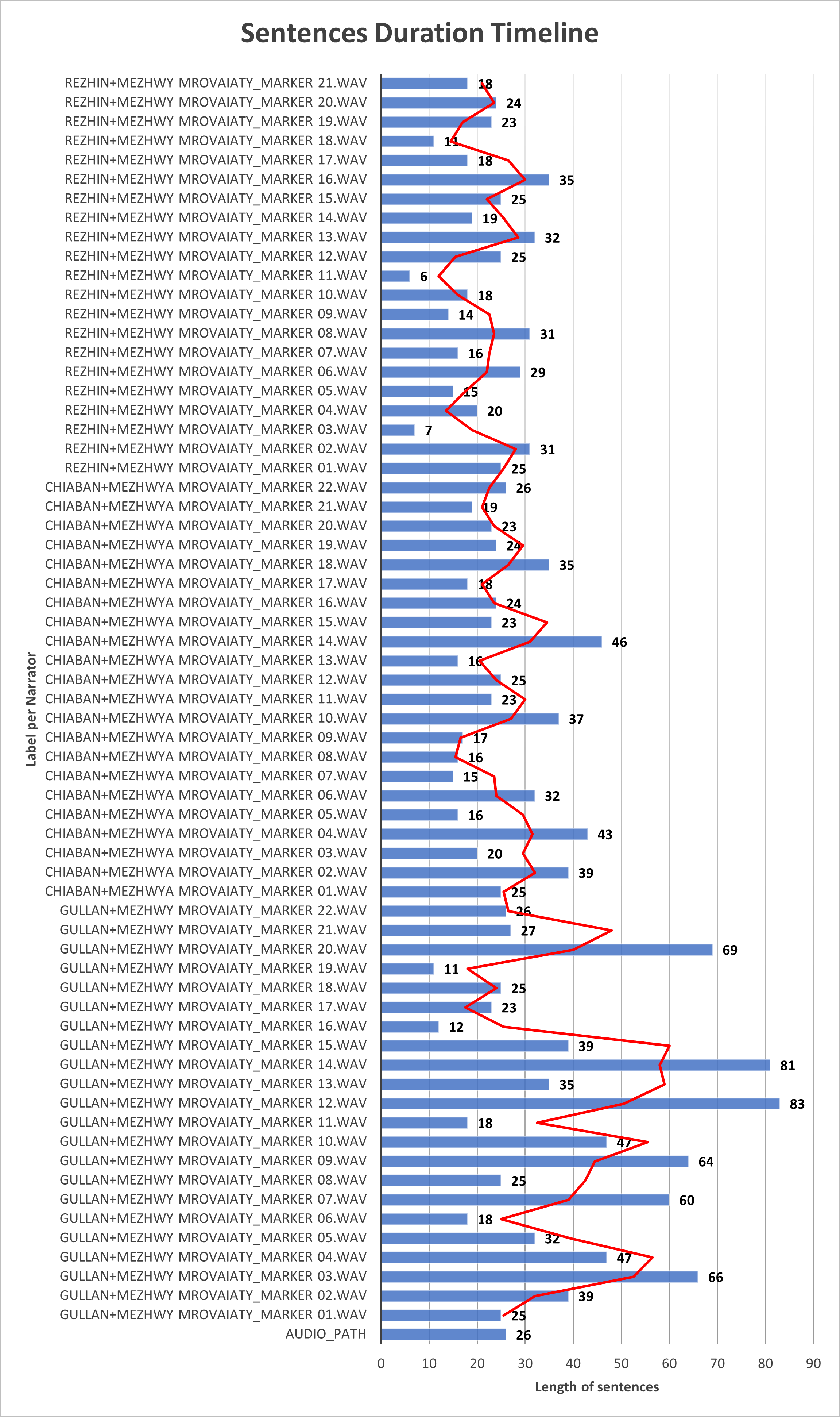}}}
    \caption{Sentences duration timeline per narrator}
    \label{Labeling-segments}
\end{figure}

The speed of reading was different among the narrators, resulting in different audio segment lengths (see Figure~\ref{Labeling-segments}). This phenomenon was the result of differences in the utterance speed of each narrator. It affected the segmentation and alignment process because we had to segment the audio at different lengths according to the narrators' utterance characteristics.    

\subsection{Fine-Tuning and Hyperparameter Configuration}

\begin{figure}[!ht]
    \centerline{\includegraphics[scale=0.07]{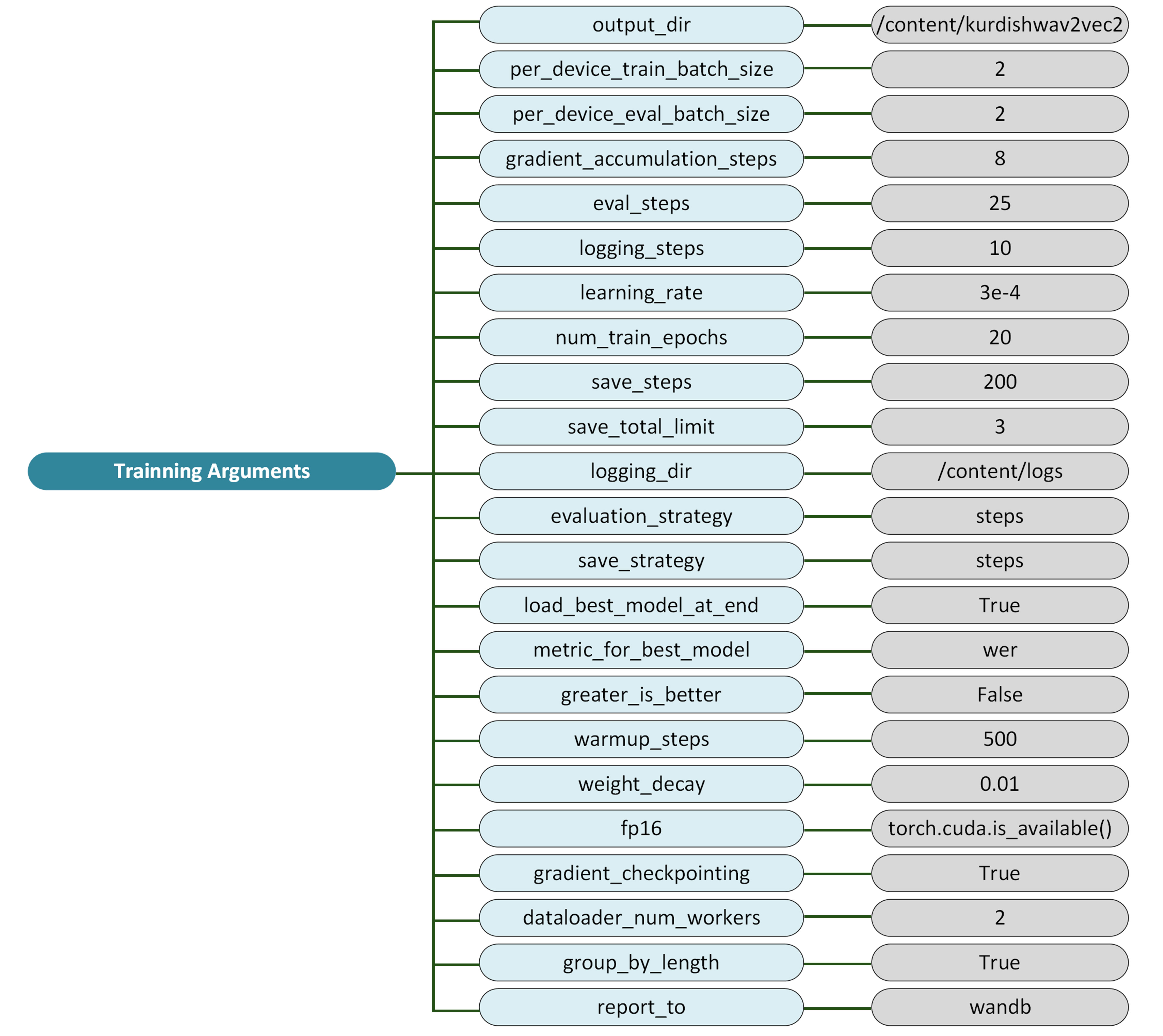}}
    \caption{Training arguments for XLSR-53}
    \label{Training arguments for XLSR-53}
\end{figure}

\begin{figure}[!ht]
    \centerline{\includegraphics[scale=0.07]{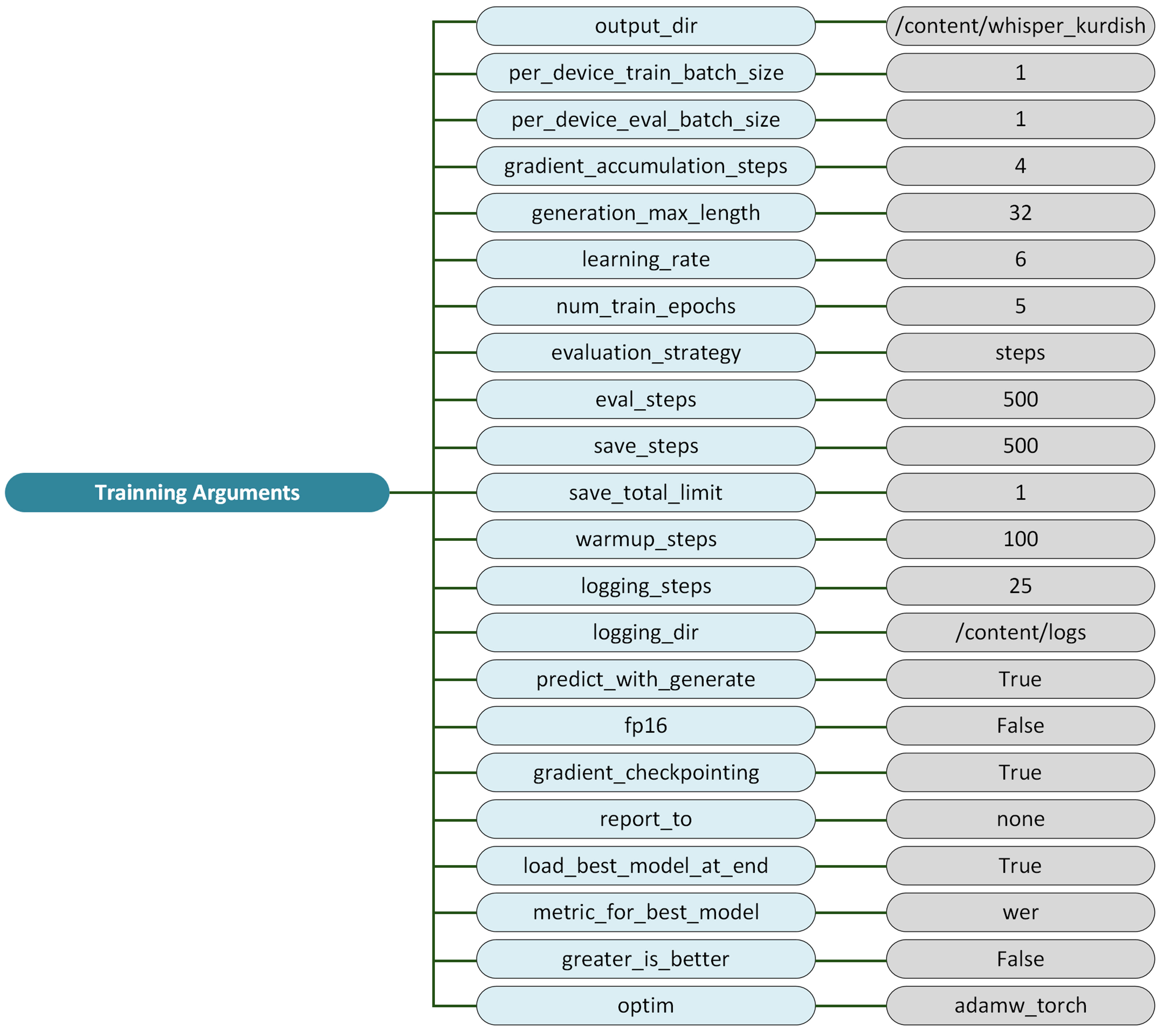}}
    \caption{Training arguments for Whisper-small}
    \label{Training-arguments-for-Whisper}
\end{figure}

Figures~\ref{Training arguments for XLSR-53} and~\ref{Training-arguments-for-Whisper} show hyperparameter configurations for Wav2Vec2-Large-XLSR-53 and Whisper-small, respectively. The configurations were set based on the available resources. For example, because of a limited GPU memory, for XSLR-53, we used a small batch size (\textit{per-device-train-batch-size = 2}) with gradient accumulation of 8 steps to give an equivalent simulation of a larger batch size. Also, since Whisper needs a large amount of memory, we set a minimum batch size of one (\textit{per-device-train-batch-size = 1}) and employed gradient accumulation across four steps to simulate a large batch size. 

\subsection{Optimization Strategy for Wav2Vec2-XLSR Training}
\begin{figure}[!ht]
    \centerline{\includegraphics[scale=0.4]{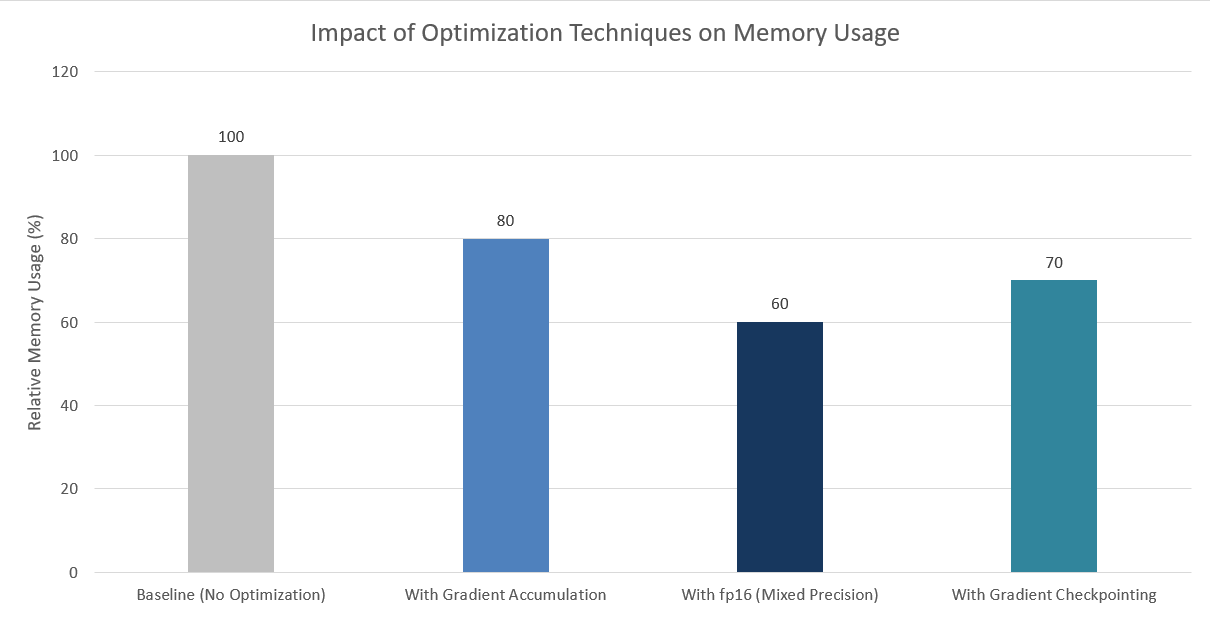}}
    \caption{Impact of Optimization Techniques on Memory Usage}
    \label{Relative Memory Usage}
\end{figure}

We use the AdamW optimizer because it is in line with standard practices in the fine-tuning of large transformer-based models, particularly the speech recognition task \cite{haswani2022methods}. As mentioned in the training arguments, the learning rate was set to \textit{3e-4}, a higher rate than the default for low-resource languages, to allow for faster convergence during the training phase. Apart from stabilizing learning and preventing overfitting, a weight decay of 0.01 was employed to regularize the model without interfering with Adam's adaptive learning rates. In addition, gradient accumulation over 8 steps was used to artificially augment the batch size, leading to improved stability under low GPU memory. Furthermore, we activated gradient checkpointing, a computationally inexpensive memory-saving method that trades off compute for memory by redoing the backpropagation and recomputing the intermediate activations. Figure~\ref{Relative Memory Usage} shows the impact of the optimization variants on memory usage. It illustrates a delicate balance between performance, memory requirements, and training efficiency, which in turn shows the possibility of proper fine-tuning under resource constraints.

\subsection{Optimization Strategy for Whisper-small Training}
Similar to Wav2Vec, we use the AdamW optimizer (specifically the adamw-torch variant) for this model, along with the weight decay from the gradient update process to avoid overfitting and preserve adaptive learning rates per parameter group, which is generally advised for Transformer-based models \cite{ferraz2023efficient}. A low learning rate of \textit{5e-6} was employed to prevent destabilizing the pretrained weights. We utilized a small batch size (1) but used gradient accumulation across 4 steps to effectively simulate the impact of a large batch size without reaching memory limits. Also, we enabled Gradient checkpointing, conserving memory by recalculating activations during backpropagation.

Unlike Wav2Vec2 training, mixed precision (fp16) was not used to maintain numerical stability as the priority during sequence generation tasks. Additionally, to enforce consistency in sequence lengths and mitigate potential decoding mismatch, a generation length limit of 32 tokens was enforced during evaluation. The best checkpoint for the model was chosen automatically based on prioritizing WER minimization, with saving and evaluation every 500 steps to enforce training consistency. These decisions were in line with conventional methods employed for low-resource STT to guarantee effective use of resources without compromising performance levels.

\subsection{Checkpoint Management Strategy During Training}

To ensure training continuity and data integrity, we established a structured checkpoint strategy that includes four checkpoints. The first checkpoint was implemented before model training to detect and resume from previously saved or incomplete checkpoints stored in the linked Drive location. The second checkpoint was implemented during the process of training to maintain synchronization with the Drive, minimizing data loss due to potential interruptions. After training was complete, the third checkpoint was established to persist the final model state so that all the remaining training data was processed and stored in the Drive. Furthermore, an additional checkpoint was used during testing of new data as a stable source by indicating the most recent saved state of the model. Before considering the discussed strategy, we faced many interruptions while training, which generated 44 checkpoints to finish the entire audio file. This organized strategy allowed for the efficient use of resources and less chance of starting training from scratch during long periods of training.

\subsection{Model Training}
\begin{figure}[!ht]
    \centerline{\includegraphics[scale=0.5]{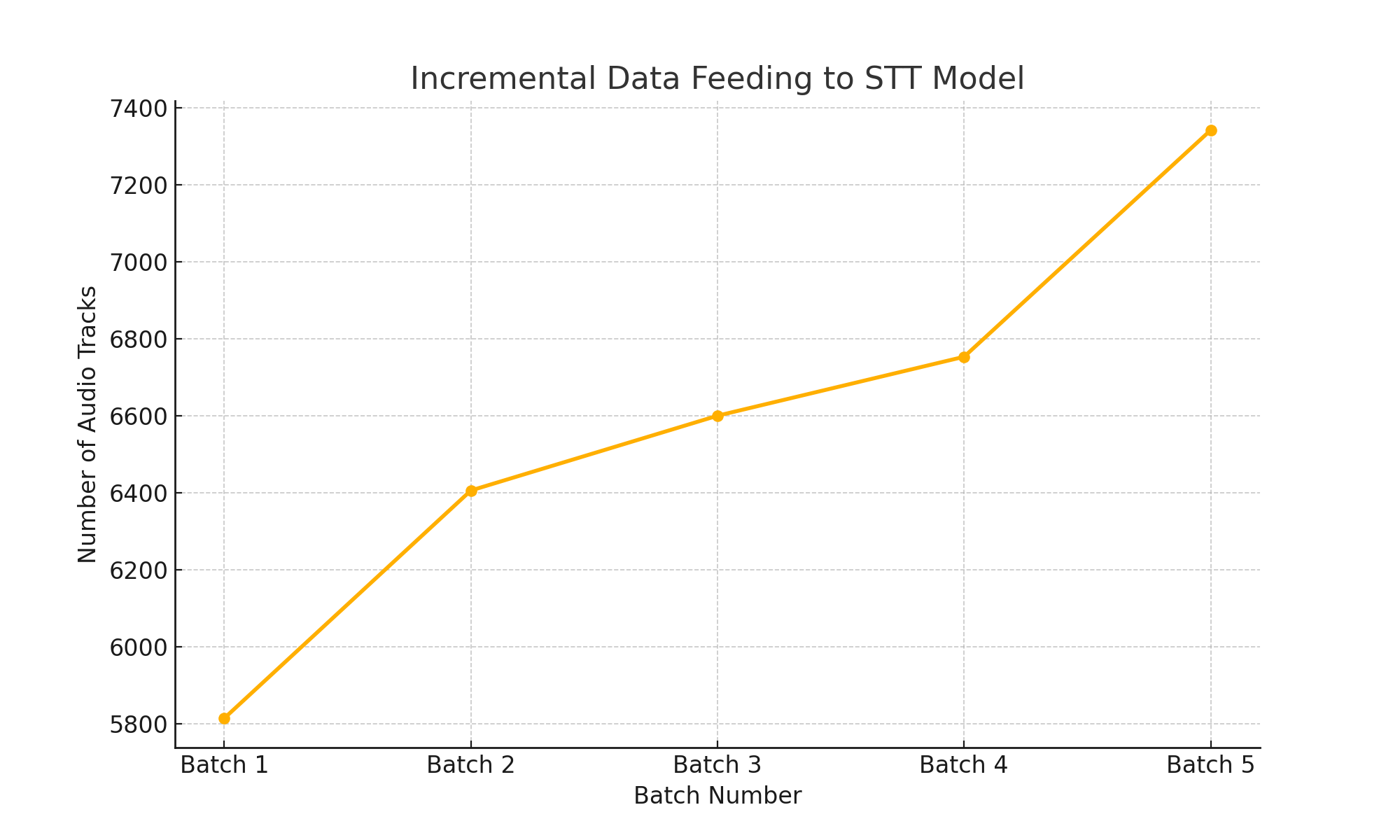}}
    \caption{Incremental training chart}
    \label{Incremental_Training_Chart.png}
\end{figure}

As mentioned, the data collection resulted in 19,193 tracks, from different channels, each paired with a short sentence transcription. This scale created challenges for model training, as feeding the entire dataset at once led to long rendering times and high resource consumption. To control and overcome this, we implemented an incremental training strategy, dividing the dataset into five smaller batches as illustrated in Figure~\ref{Incremental_Training_Chart.png}. This allowed for more efficient resource management, it helped us to prevent overfitting, and reduced training interruptions. Our approach in this stage aligns with the findings of \cite{vandeven2022incremental} researchers, who highlight the effectiveness of (incremental/continual) learning techniques while working with non-stationary data streams. Their research work demonstrates that incremental strategies can enhance performance and stability when training deep learning models on large, evolving datasets.

\subsection{Wav2Vec Training}
\begin{figure}[!ht]
    \centerline{\includegraphics[scale=0.4]{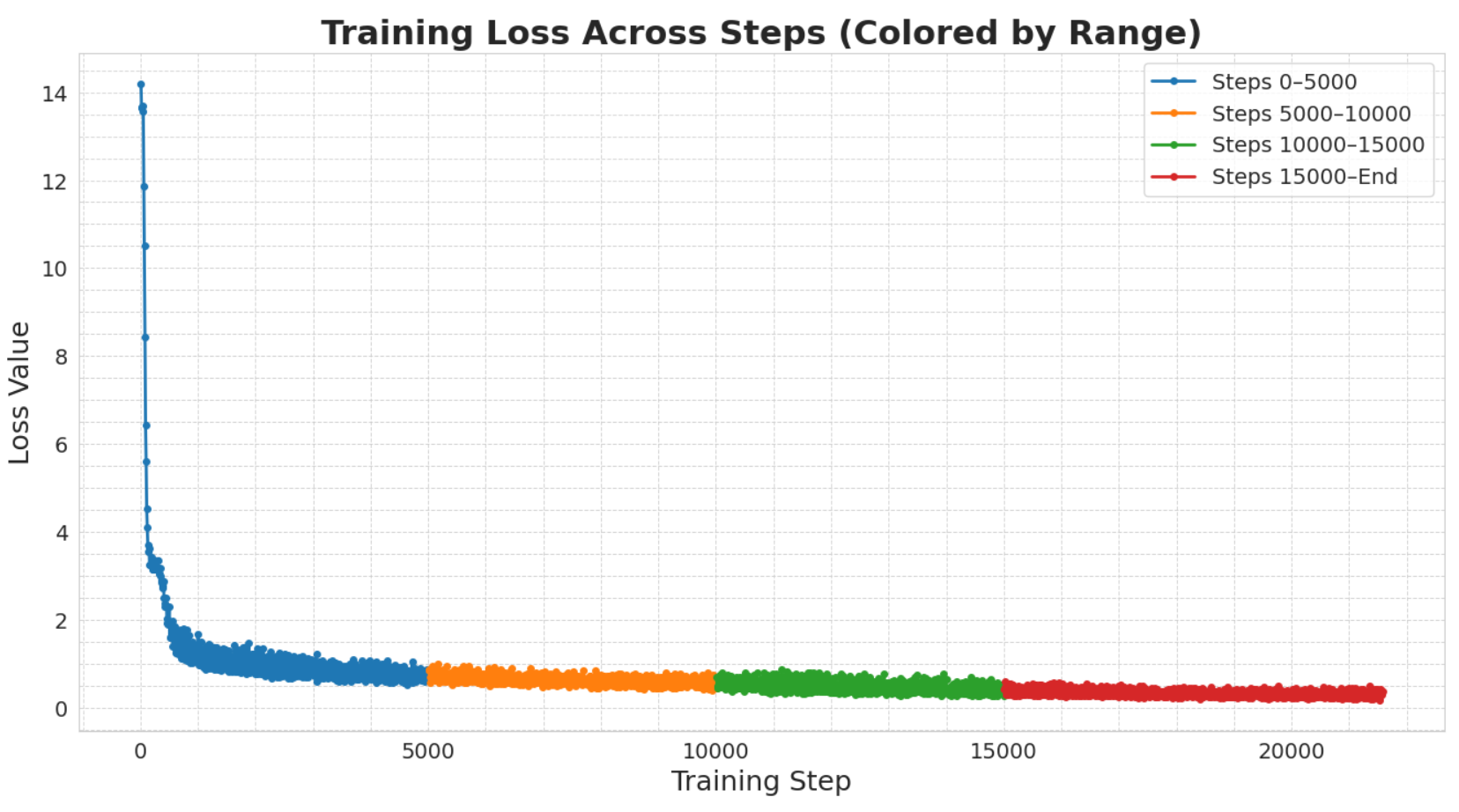}}
    \caption{The total of Training Loss from different ranges of steps}
    \label{VLandTL-XLSR}
\end{figure}

\begin{figure}[!ht]
    \centerline{\includegraphics[scale=0.4]{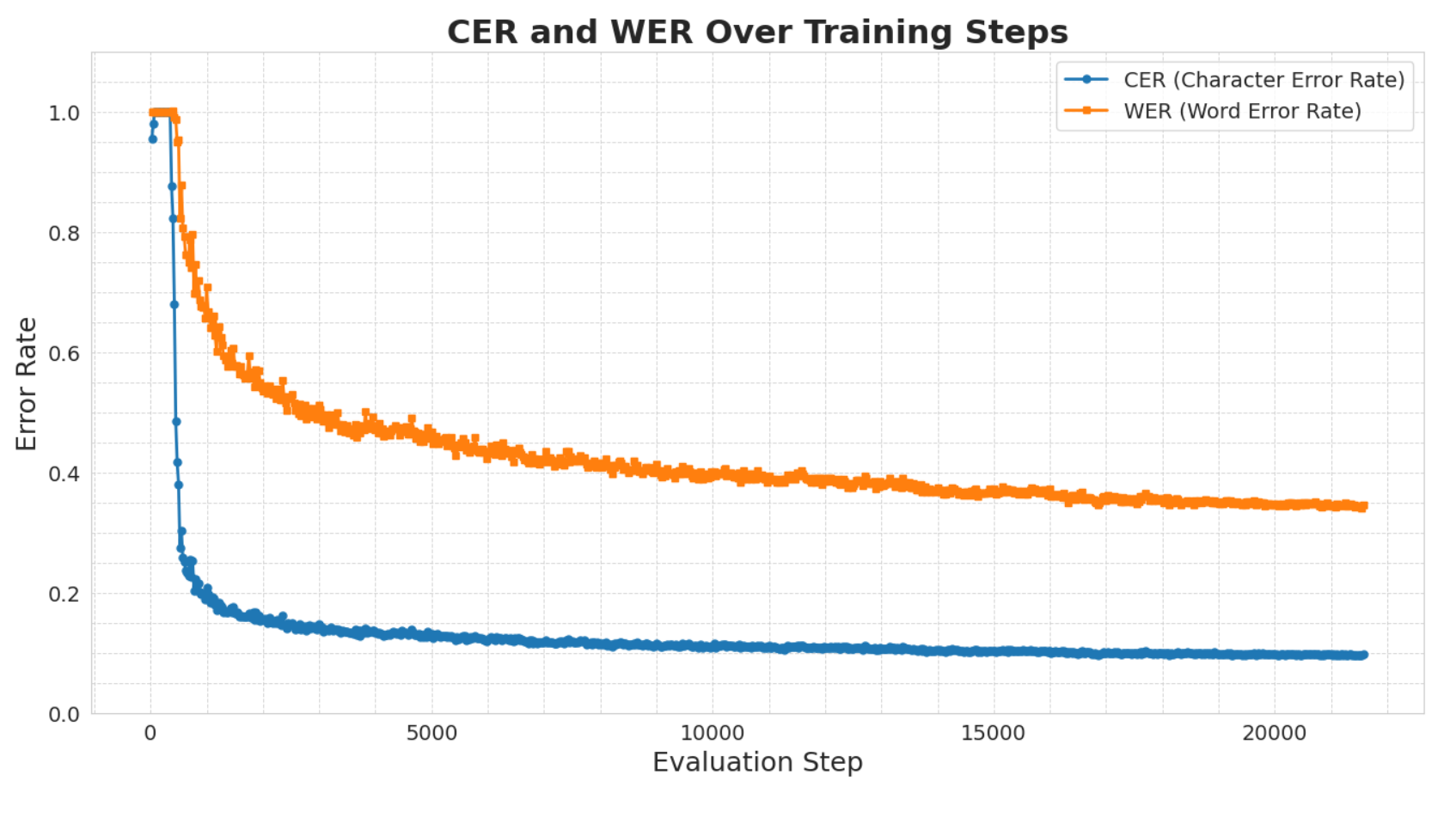}}
    \caption{WER and CER for Wav2Vec2 model training result}
    \label{WERandCER-XLSR}
\end{figure}

We used 90\% of the data for training and the rest for testing, resulting in 17,272 training examples and 1,920 testing examples. This scheme serves better due to the size of the speech data, which is considered small for STT tasks. The model was trained for a total of 21,581 steps, covering 20 epochs. The training script steps were executed sequentially to ensure proper tracking to avoid known errors. As Figures~\ref{VLandTL-XLSR} and~\ref{WERandCER-XLSR} illustrate, a progressive reduction in both training and validation loss happened throughout the process. Table~\ref{Table:Wav2Vec2_evaluation} presents a summary of the metrics for the entire process.

\begin{table}[!ht]
\centering
\caption{Results for Wav2Vec}
\label{Table:Wav2Vec2_evaluation}
\begin{tabular}{|l|l|} 
    \hline
    \cellcolor{lightgray}\textbf{Metric} & 
    \cellcolor{lightgray}\textbf{Value} \\
    \hline
     WER (Start) &  100\% \\ 
    \hline
     WER (End) &  34.55\% \\ 
    \hline
     Best WER &  34.31\% \\ 
    \hline
     CER (Start) &  95.52\% \\ 
    \hline
     CER (End) &  9.62\% \\ 
    \hline
     Best CER &  9.58\% \\ 
    \hline
     Validation Loss (Start) &  14.04 \\ 
    \hline
     Validation Loss (End) &  0.4587 \\ 
    \hline
     Training Loss (Start) &  12.62\% \\ 
    \hline
     Training Loss (End) &  0.30\% \\ 
    \hline
     WER Accuracy &  65.45\% \\ 
    \hline
     CER Accuracy &  90.38\% \\ 
    \hline
\end{tabular}
\end{table}


\subsubsection{Whisper Training}

\begin{figure}[!ht]
    \centerline{\includegraphics[scale=0.4]{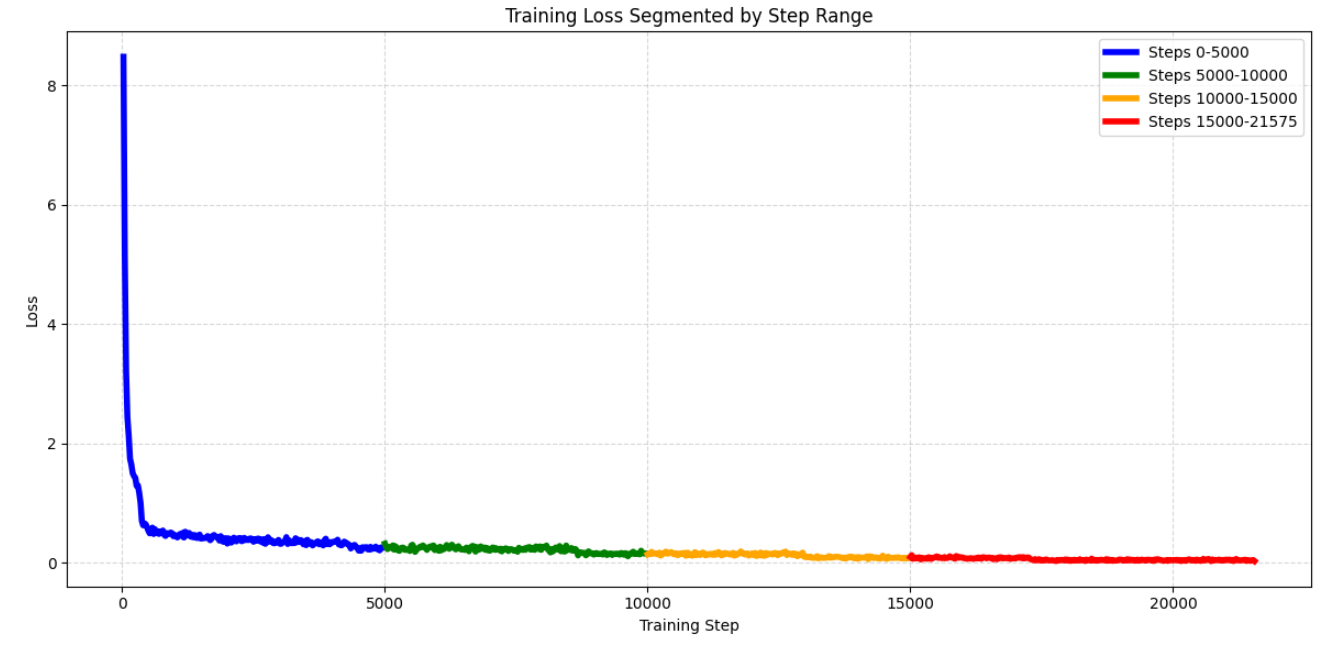}}
    \caption{The total of Training Loss from different ranges of steps for Whisper-small}
    \label{VLandTL-WS}
\end{figure}

\begin{figure}[!ht]
    \centerline{\includegraphics[scale=0.4]{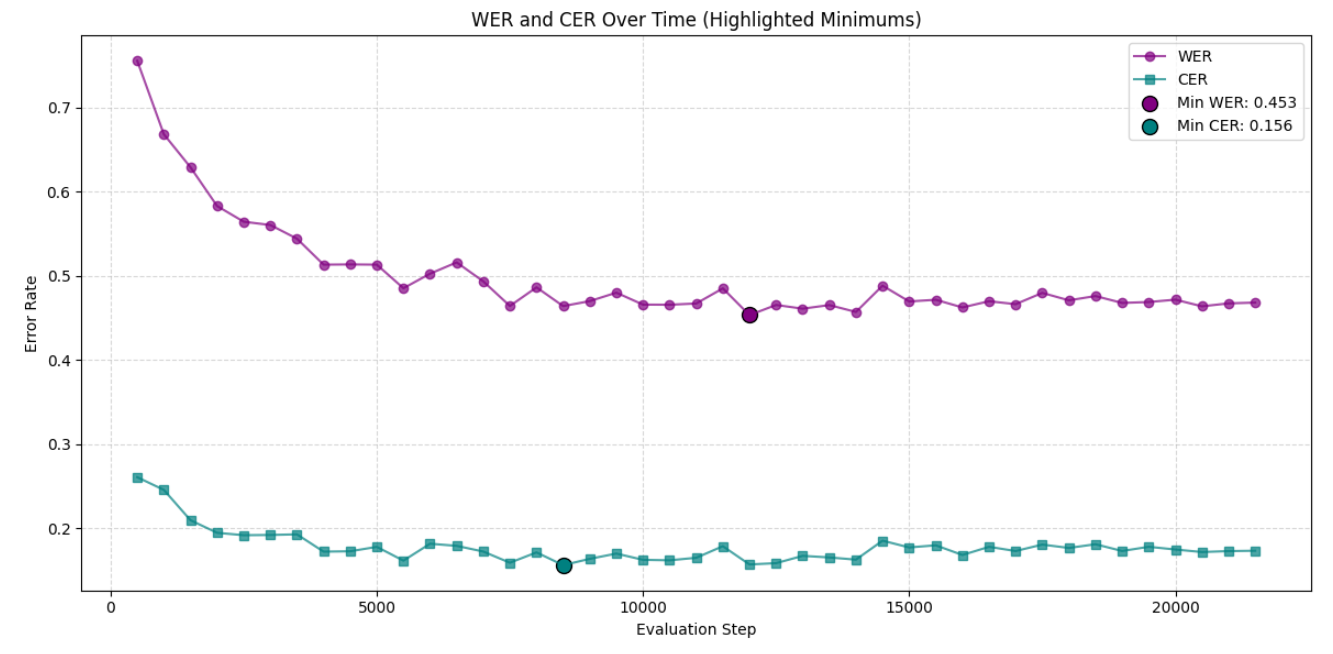}}
    \caption{WER and CER for Whisper-small model training result}
    \label{WERandCER-WS}
\end{figure}

The Whisper training took a total of 21,500 steps. Similar to Wav2Vec training, we observed a consistent reduction in both training and validation loss. Figure~\ref{VLandTL-WS} illustrates the training loss, and Figure~\ref{WERandCER-WS} shows the WER and CER during the training process. Table~\ref{tab:Whisper_small_evaluation} provides a summary of the metrics for the entire process.  

\begin{table}[!ht]
\centering
\caption{Results for Whisper}
\label{tab:Whisper_small_evaluation}
\begin{tabular}{|l|l|} 
    \hline
    \cellcolor{lightgray}\textbf{Metric} & 
    \cellcolor{lightgray}\textbf{Value} \\
    \hline
     WER (Start) &  75.57\% \\ 
    \hline
     WER (End) &  46.83\% \\ 
    \hline
     Best WER &  45.33\% \\ 
    \hline
     CER (Start) &  26.07\% \\ 
    \hline
     CER (End) &  17.33\% \\ 
    \hline
     Best CER &  15.61\% \\ 
    \hline
     Validation Loss (Start) &  0.5886 \\ 
    \hline
     Validation Loss (End) &  0.3250 \\ 
    \hline
     Training Loss (Start) &  8.479\% \\ 
    \hline
     Training Loss (End) &  0.0419\% \\ 
    \hline
     WER Accuracy &  53.17\% \\ 
    \hline
     CER Accuracy &  82.67\% \\ 
    \hline
\end{tabular}
\end{table}

\subsection{Evaluation}
The Wav2Vec model was loaded from its final checkpoint. The audio files were resampled to 16 kHz to match the model's expected input configuration. We initially tested the model on a manually selected sample file, consisting of 30 new audio tracks, and verified the transcription quality as shown in Table~\ref{Table:UnseenAudio}. Ultimately, the achieved transcriptions of the test audio samples were exported and stored in a CSV file under the name(test-results.csv) in the same location. The testing process revealed that the trained model could generalize to new audio inputs outside the training dataset and produced very accurate transcriptions.

\begin{table}[htb]
  \centering
  \caption{Samples of Unseen Audio Segments}
  \label{Table:UnseenAudio}
  \begin{tabular}{|m{4.5cm}|>{\raggedleft\arraybackslash}b{10cm}|} 
    \hline 
    \rowcolor[gray]{0.93} 
    \centering\footnotesize\textbf{Unseen Audio Segment} & 
    \footnotesize\textbf{Narrated Text} \\ 
    \hline 
    \small\texttt{chiaban+Mezhw\_Marker 01.wav} & {\small \RL{چیرۆکا مێژوویا مرۆڤایه‌تیێ}} \\ 
    \hline 
    \small\texttt{chiaban+Mezhw\_Marker 02.wav} & {\small \RL{پاشای گوتە شیرەتکارێ خو: }} \\ 
    \hline 
    \small\texttt{chiaban+Mezhw\_Marker 03.wav} & {\small \RL{من دڤێت ئەز مێژوویا مرۆڤایەتیێ بزانم!}} \\ 
    \hline 
    \small\texttt{vahin+Mezhw\_Marker 04.wav} & {\small \RL{شیرەتکارێ وی ژی چوو و پشتی دەھ سالان ل مێژوویا مرۆڤایەتیێ گەریان، }} \\ 
    \hline 
    \small\texttt{vahin+Mezhw\_Marker 05.wav} & {\small \RL{هاتە دەف پاشایی دەھ بەرگێن مێژوویێ یێن نڤێساین }} \\ 
    \hline 
    \small\texttt{vahin+Mezhw\_Marker 06.wav} & {\small \RL{وگلهـ خو ئینا و هاتە دەف پاشایی،}} \\ 
    \hline 
  \end{tabular}
\end{table}


Similarly, we tested Whisper Small to see the model's results; a test session was run on a new batch of 30 previously untrained Kurdish Badini audio samples. The audio files were resampled at 16 kHz to align with the model's input requirements. The lengthy testing procedure was automated with the trained model and tokenizer, and the produced transcriptions were exported into a CSV file called Whisper\_test\_results.csv, as shown in Table~\ref{Table:Wav2VecTranscriptionsresult}.

\begin{table}[!ht]
\begin{threeparttable}
    \centering
    \caption{Wav2Vec2 Test Results - Audio File Transcriptions}
    \label{Table:Wav2VecTranscriptionsresult}
    \begin{tabular}{|m{4.5cm}|>{\raggedleft\arraybackslash}b{10cm}|} 
        \hline 
        \rowcolor[gray]{0.93} 
        \footnotesize\textbf{Audio File} & \footnotesize\textbf{Wav2Vec2 Transcription} \\ 
     \hline 
    \small\texttt{chiaban+Mezhw\_Marker 01.wav} & {\small \RL{چیرۆکا مێژوویا مروڤایەتیێ}} \\ 
    \hline 
    \small\texttt{chiaban+Mezhw\_Marker 02.wav} & {\small \RL{پاشای گوتە شیرەتکارێ خو: }} \\ 
    \hline 
    \small\texttt{chiaban+Mezhw\_Marker 03.wav} & \textcolor{red}{\small \RL{من دڤێت ئەز مێژوویا مرۆڤایهتیێ بزانم،!}} \\ 
    \hline 
    \small\texttt{vahin+Mezhw\_Marker 04.wav} & \textcolor{red}{\small \RL{پشتی دەھ سالان ل مێژوویا مرۆڤایه‌تیێ دگەریان، }} \\ 
    \hline 
    \small\texttt{vahin+Mezhw\_Marker 05.wav} & \textcolor{red}{\small \RL{دەھ بەرگێن مێژووی یێ نڤێساین و }} \\ 
    \hline 
    \small\texttt{vahin+Mezhw\_Marker 06.wav} & \textcolor{red}{\small \RL{گەل خو ئینا و،}} \\ 
    \hline 
    \end{tabular}
    \begin{tablenotes}
        \item[]Note: The entries in red are incorrect or incomplete transcriptions.
    \end{tablenotes}
\end{threeparttable}
\end{table}
\begin{table}[!ht]
    \centering
    \caption{Whisper Test Results - Audio File Transcriptions}
    \label{Table:WhisperTranscriptionsresult}
    \begin{tabular}{|m{4.5cm}|>{\raggedleft\arraybackslash}b{10cm}|} 
        \hline 
        \rowcolor[gray]{0.93} 
        \footnotesize\textbf{Audio File} & \footnotesize\textbf{Whisper Transcription} \\ 
  \hline 
    \small\texttt{chiaban+Mezhw\_Marker 01.wav} & {\small  {\RL{چیرۆکا {مێژوویا} مرۆڤایه‌تیێ}}} \\ 
    \hline 
    \small\texttt{chiaban+Mezhw\_Marker 02.wav} & {\small \RL{پاشای گوتە شیرەتکارێ خو: }} \\ 
    \hline 
    \small\texttt{chiaban+Mezhw\_Marker 03.wav} & {\small \RL{من دڤێت ئەز مێژوویا مرۆڤایەتیێ بزانم!}} \\ 
    \hline 
    \small\texttt{vahin+Mezhw\_Marker 04.wav} & {\small \RL{شیرەتکارێ وی ژی چوو و پشتی دەھ سالان ل مێژوویا مرۆڤایەتیێ گەریان، }} \\ 
    \hline 
    \small\texttt{vahin+Mezhw\_Marker 05.wav} & {\small \RL{هاتە دەف پاشایی دەھ بەرگێن مێژوویێ یێن نڤێساین }} \\ 
    \hline 
    \small\texttt{vahin+Mezhw\_Marker 06.wav} & {\small \RL{وگلهـ خو ئینا و هاتە دەف پاشایی،}} \\ 
    \hline 

    \end{tabular}
\end{table}

Since no Badini STT was available, we compared the obtained results with an STT for another Kurdish dialect, the work by \newcite{Abdullah2023BreakingWalls} who developed an end-to-end STT for Central Kurdish (Sorani), using a 100-hour speech corpus, collected from diverse sources, achieving a WER of 11.8\%. Similarly, \newcite{Das2021MultiDialect} tackled multi-dialect English ASR using 60,000 hours of speech data and achieved a WER of 4.74\% using ensemble-based attention mechanisms. That reaffirms the bottleneck of NLP tasks, which is the amount of available data. However, the results also showed that with a considerably small amount of data, workable models could be developed to pave the way for further enhancements. 

We also noticed a situation we did not expect at the beginning of the project. Our expectation was that the training of the models based on native narrations could create a more accurate model. However, despite their limited fluency, the non-native speakers tended to produce a better narration quality and more explicit articulation of vowel sounds within sentences. That was beneficial for model training, as native speakers tend to rely on prosodic familiarity and, unintentionally, skip vowel sounds because of natural speech rhythm and internalized phonological stresses. But non-native speakers read more intentionally, which enhances phonetic clarity and produces fewer skipped vowels inside the words.

In summary, the Wav2Vec2 model outperformed the Whisper-small model in nearly all evaluation metrics. It achieved a higher degree of alignment with the Badini dataset, lower WER and CER values, and demonstrated a more stable and effective training trajectory. In our case, the Wav2Vec2 model proved to be a more suitable and accurate solution. 

\section{Conclusion}
We collected over 17 hours of speech data from story narrations by native and non-native speakers in Badini Kurdish and extracted over 19193 segments of 1 to 12 seconds, covering 1 to 8 words. Two Speech to Text (STT) frameworks, Wav2Vec2 and Whisper Small, were evaluated and compared based on Word Error Rate (WER) and Character Error Rate (CER). The Wav2Vec2 model generally outperformed Whisper in both WER and CER. However, both models demonstrated the potential to be adapted for an STT task with limited data. The results contribute to the enrichment of resources for Kurdish as a low-resource language.

In the future, we are interested in expanding the dataset and including a wider range of narrators to cover more inclusive accents from other Badini-speaking cities and areas. Also, through a web-based application, we aim to increase the speech data to cover more diverse topics and retrain the models periodically to improve their accuracy. We are also interested in an automatic correction module that is set to be incorporated to correct frequent errors in the transcription.

\section*{Data Availability and Companion Website}
To test the model performance, we are in the process of developing a website. The interested readers can regularly check \url{https://bkstt.com/} to follow the development progress.
The data will be made publicly available at \url{https://kurdishblark.github.io/} in the future.

\section*{Ethical Consideration}
The author of the books provided his consent that the collected data could be shared for research purposes. All narrators also provided their consent for using their recorded narrations in this research. 

\section*{Acknowledgment}

We appreciate the support of all those who have contributed to this project. Our special gratitude goes to MR. Nihad Sa'adullah for providing the books in Badini as essential rough data. We extend our warm appreciation to Gullan Tahsin, Chyaban Junaid, Rezhin Yousif, Hevi Esa, Vahin Hasan, and Ahmad Omed for narrating the books.

\bibliographystyle{lrec}

\bibliography{BKSTT}

\newpage
\appendix
\setcounter{page}{1}
\setcounter{figure}{0}  
\renewcommand{\thefigure}{A.\arabic{figure}} 
\renewcommand{\thepage}{A\arabic{page}}
\renewcommand{\thesection}{A\arabic{section}}
\renewcommand{\thesubsection}{A\arabic{section}.\arabic{subsection}}	
\section*{Appendix A\\ Images Referred to in the Method Section}

\begin{figure}[!ht]
    \centerline{\includegraphics[scale=0.4]{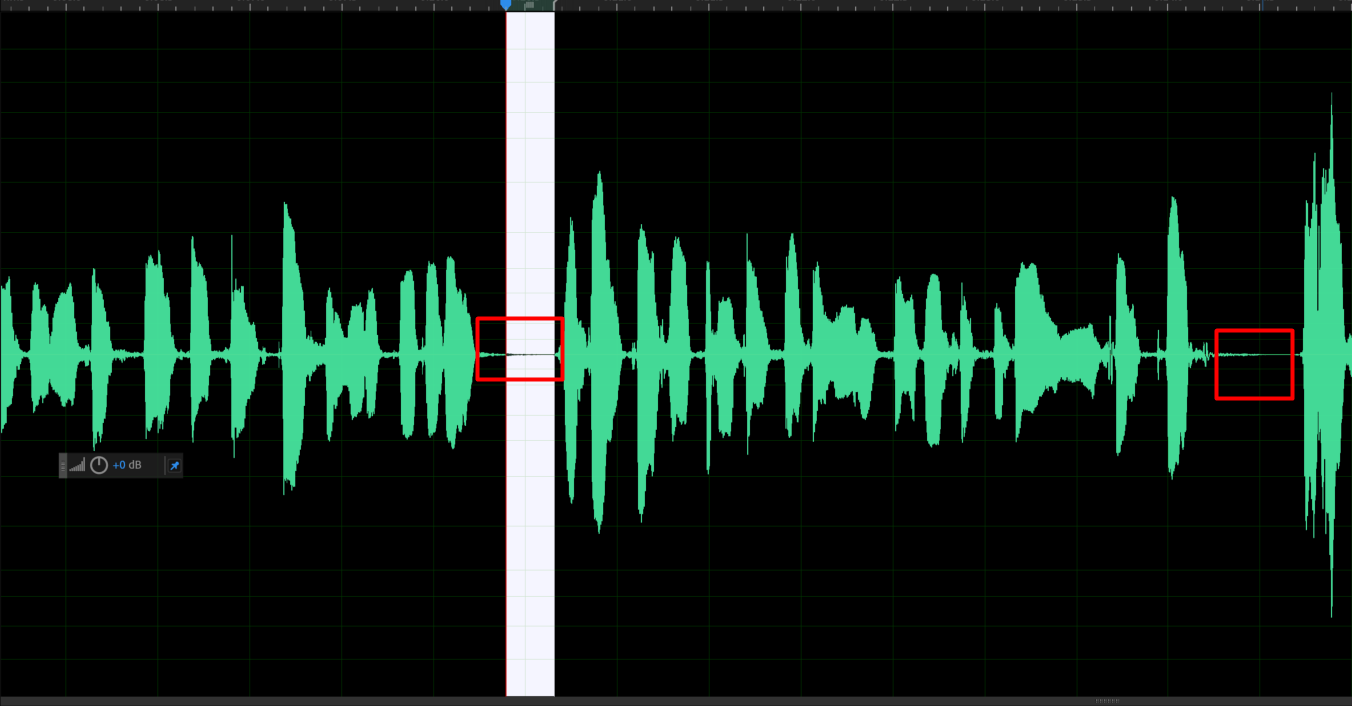}}
    \caption{Completed noise reduction}
    \label{zeronoise}
\end{figure}

\begin{figure}[!ht]
    \centerline{\includegraphics[scale=0.36]{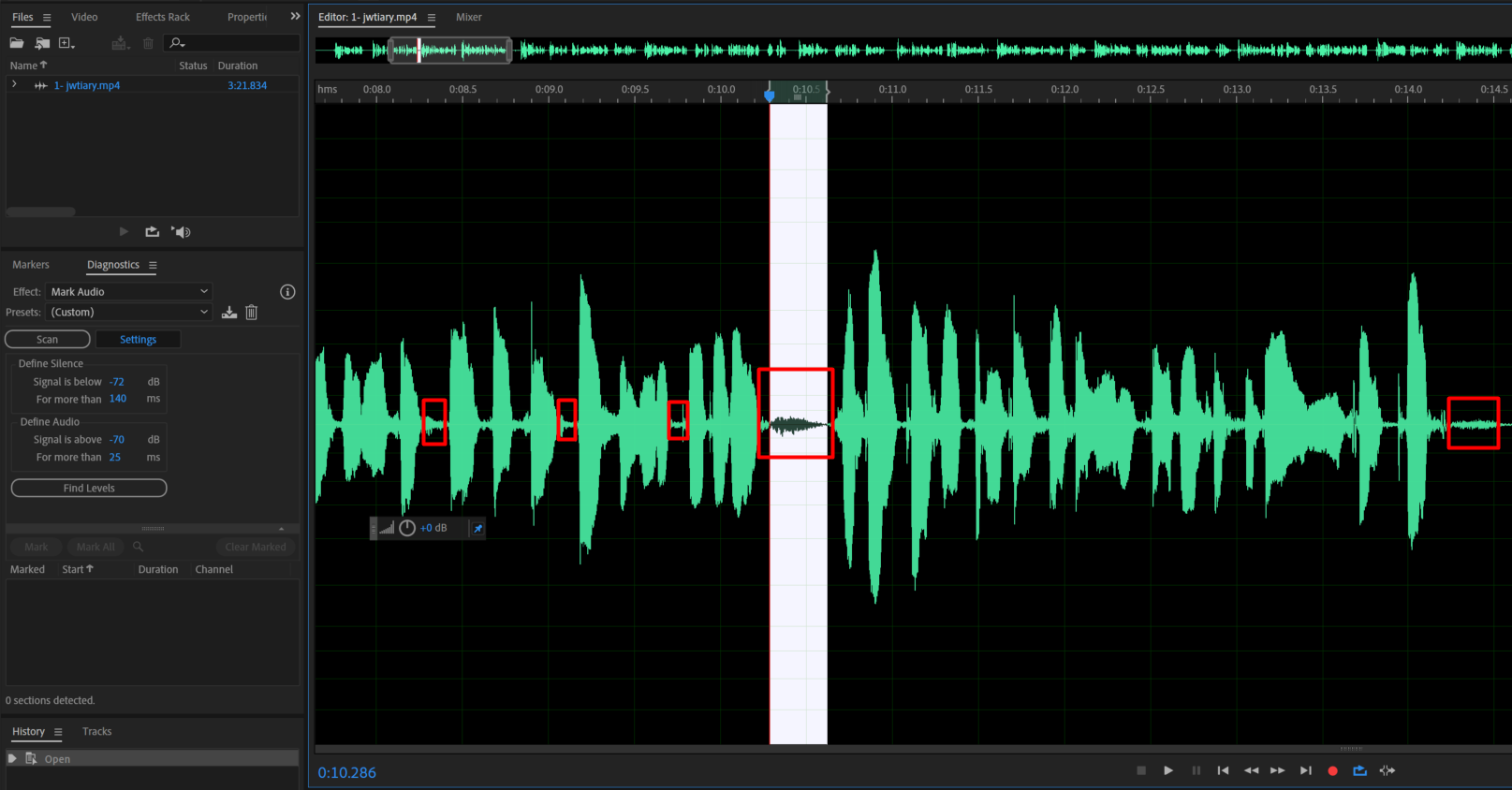}}
    \caption{A sample of noise capturing}
    \label{capturenoise}
\end{figure}

\begin{figure}[!ht]
    \centerline{\includegraphics[scale=0.9]{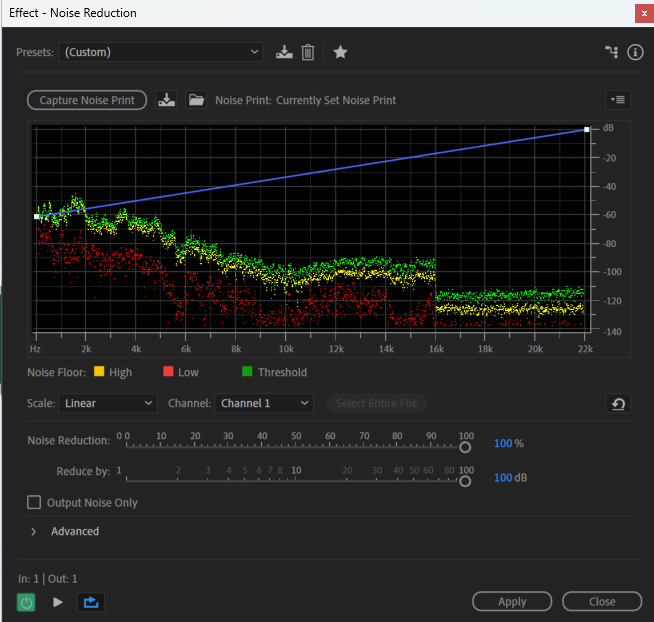}}
    \caption{Splitting the noise from the actual voice}
    \label{NoiseReduction}
\end{figure}

\begin{figure}[!ht]
    \centerline{\includegraphics[scale=0.3]{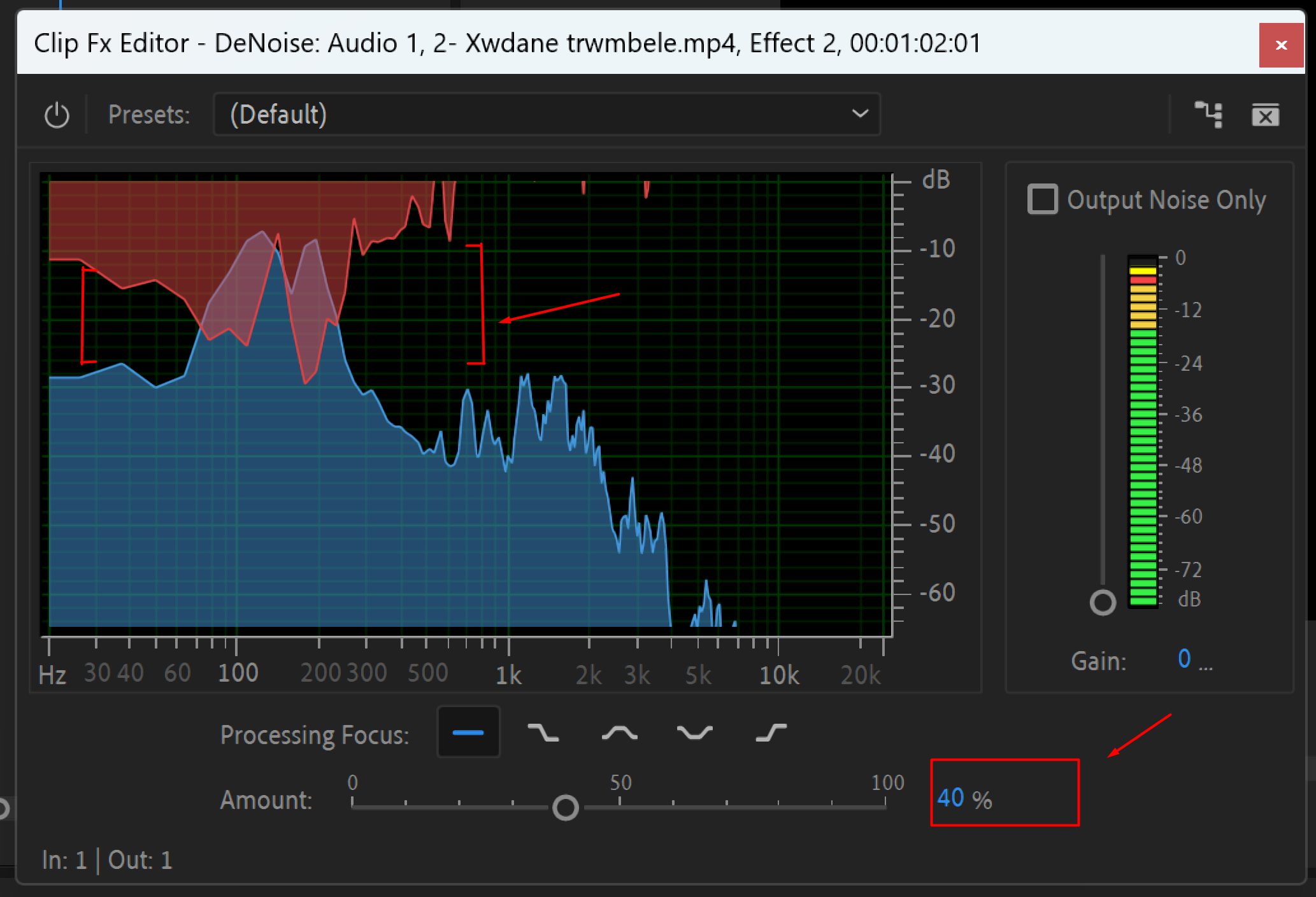}}
    \caption{Before applying the enhancer voice filter by 11 percent}
    \label{before-noise}
\end{figure}

\begin{figure}[!ht]
    \centerline{\includegraphics[scale=0.3]{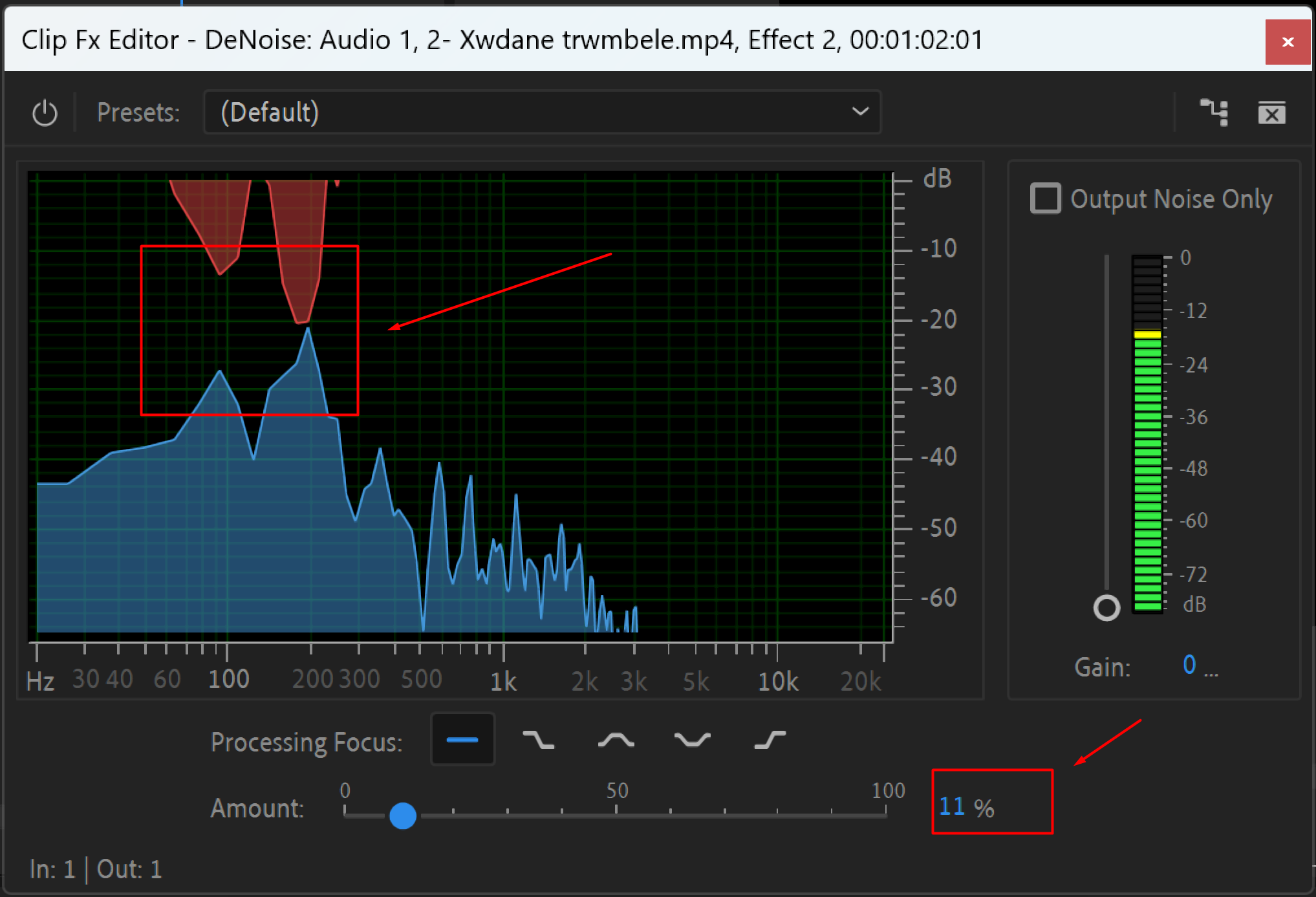}}
    \caption{After applying the enhancer voice filter by 11 percent}
    \label{after-noise}
\end{figure}

\begin{figure}[!ht]
    \centerline{\includegraphics[scale=1.5]{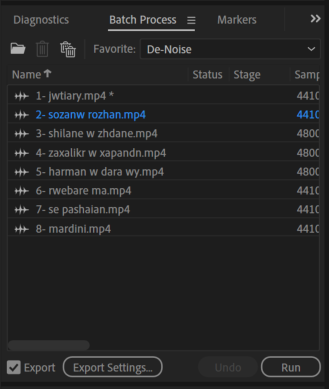}}
    \caption{Batch processing Panel}
    \label{batch-processing}
\end{figure}

\begin{figure}[!ht]
    \centerline{\includegraphics[scale=1.2]{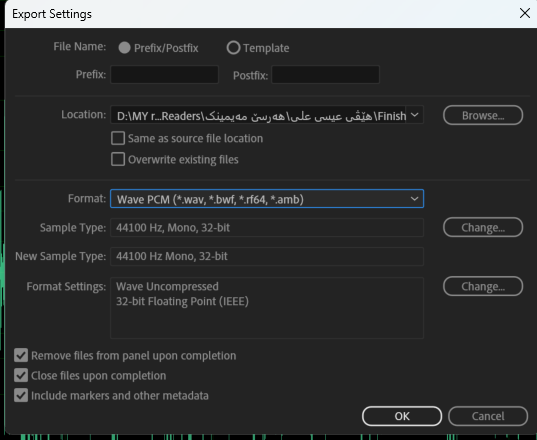}}
    \caption{Export format setting in Batch processing Panel}
    \label{Export_option}
\end{figure}

\begin{figure}[!ht]
    \centerline{\includegraphics[scale=0.5]{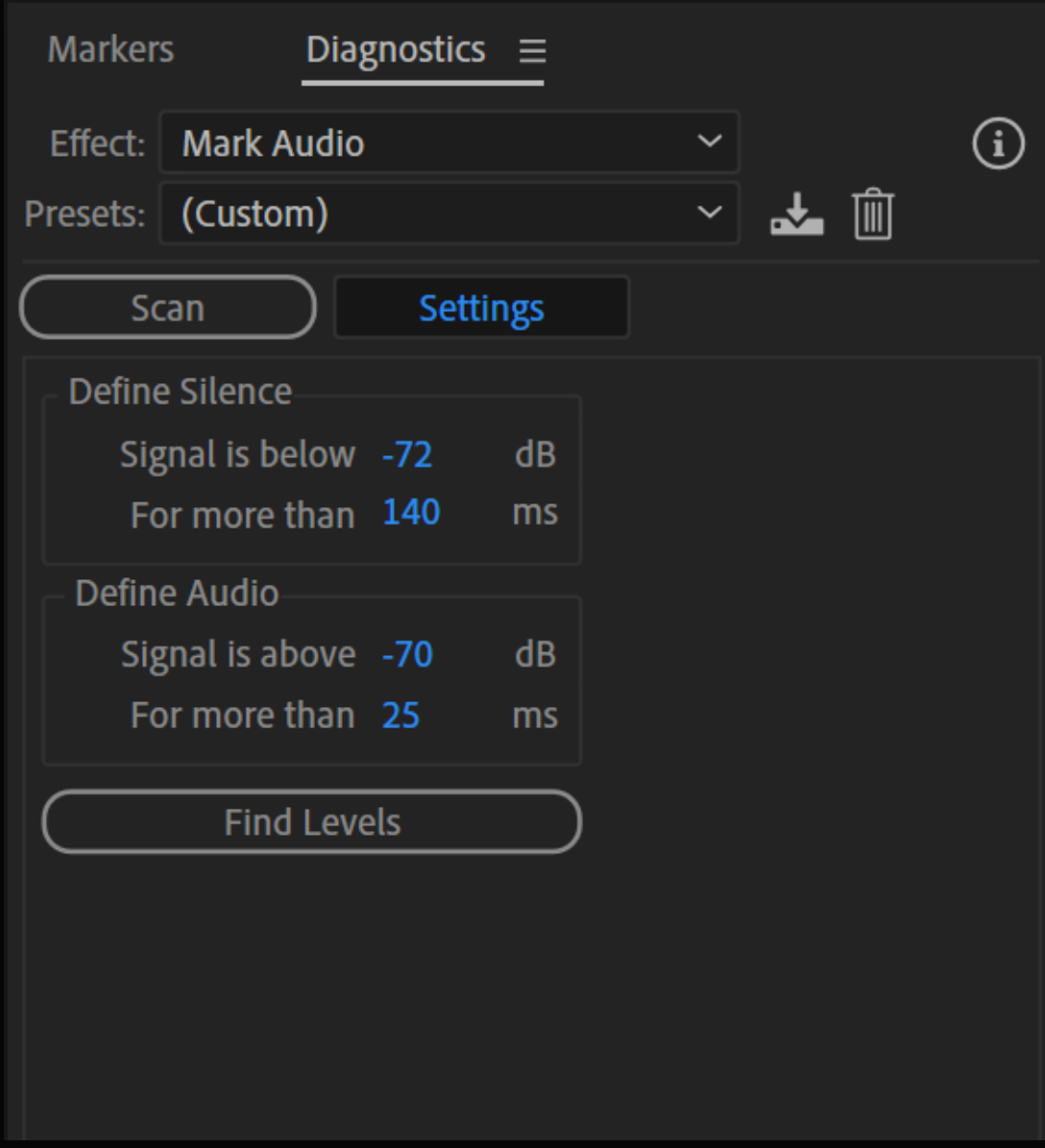}}
    \caption{Diagnostics Panel: The "Mark Audio" effect provides identification and segmentation sections of silence and active audio based on customizable dB and time thresholds.}
    \label{diagnostics}
\end{figure}
\begin{figure}[t!]
    \centerline{\includegraphics[scale=0.40]{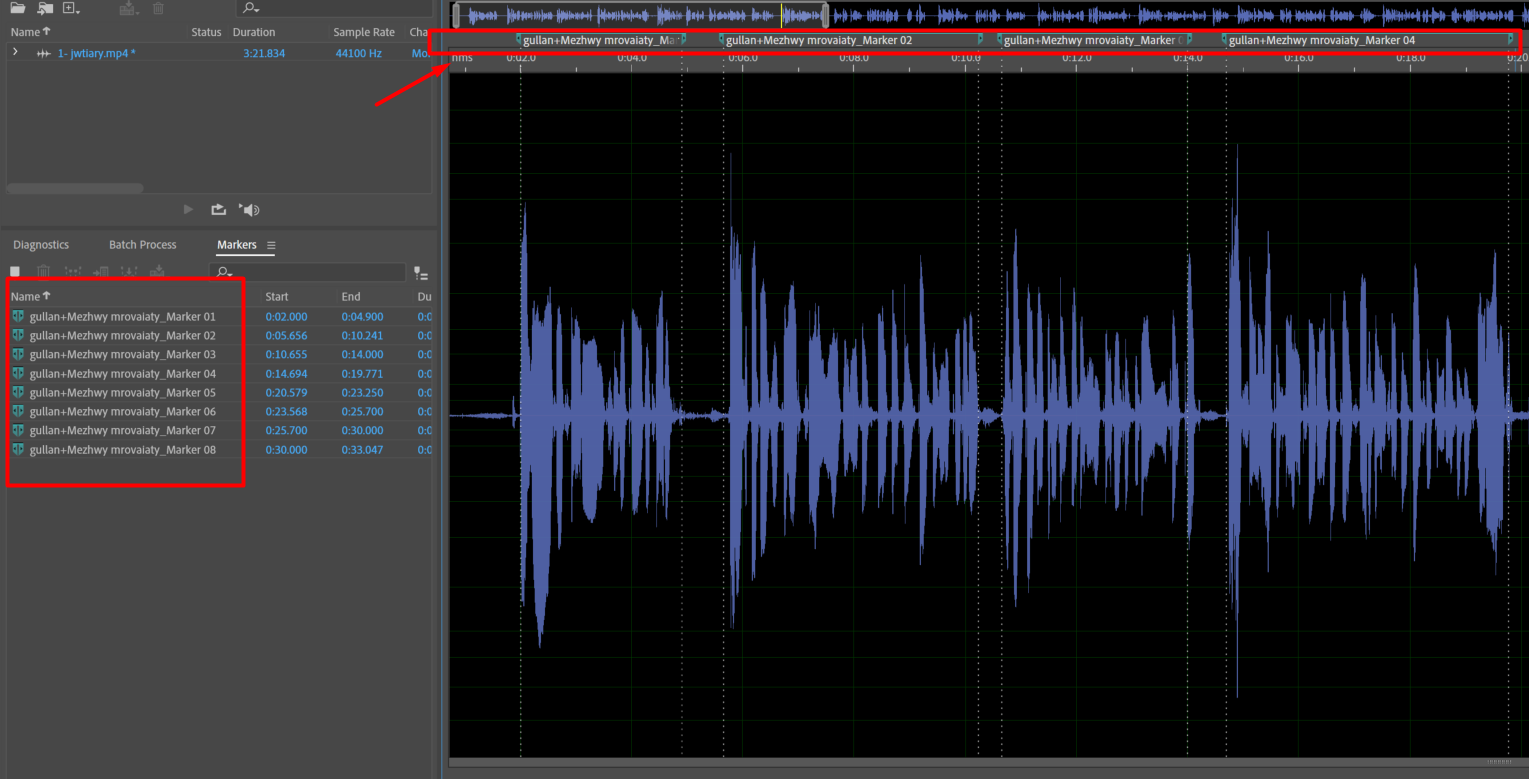}}
    \caption{Labeling the spoken sentence}
    \label{labling_names}
\end{figure}

\end{document}